\documentclass[a4paper]{svproc}
\usepackage{url}

\usepackage{graphicx}
\usepackage{graphics} 
\usepackage{epsfig} 
\usepackage{wrapfig}
\usepackage{times} 
\usepackage{amsmath} 
\usepackage{amssymb}  
\usepackage{hyperref}
\usepackage{booktabs}
\usepackage{multirow}
\newcommand{\thickhline}{\specialrule{0.8pt}{0pt}{0pt}}
\newcommand{\robotname}{Olympus }
\usepackage[caption=false]{subfig}
\usepackage{siunitx}
\begin{document}
\mainmatter              
\title{Towards Low-Gravity Planetary Exploration using Reinforcement Learning for Walking, Jumping, and In-flight Attitude Control}
\titlerunning{Planetary Exploration using Reinforcement Learning}  
%
\author{Jørgen Anker Olsen and Kostas Alexis}
\authorrunning{Olsen et al.} 
%
%
\institute{The authors are with the Autonomous Robots Lab, NTNU, O.S. Bragstads Plass 2D, 7034, Trondheim, Norway.
\email{jorgen.a.olsen@gmail.com}, \\ home page:
\texttt{https://ntnu-arl.github.io/olympus/}
}

\maketitle              

\begin{abstract}
This paper presents reinforcement learning (RL) policies for dynamic quadrupedal locomotion in planetary exploration scenarios. Building on a task-optimized quadruped with a 5-bar leg design, we develop RL policies for walking, vertical jumping, forward jumping, and in-flight attitude control, explicitly tailored to the reduced gravity on Mars. These policies jointly enable such robots to overcome obstacles larger than themselves through coordinated jumping and precise in-flight reorientation for safe landings. We demonstrate Sim2Real transfer of the attitude control policy on the \robotname quadruped through single-axis reorientation tests, while all locomotion policies are validated in simulation. A complete Mars exploration mission scenario demonstrates coordinated policy deployment across challenging terrain. Experimental results show \ang{90} attitude reorientation in 2.6 seconds, with simulations demonstrating 3.1 meter vertical jumps and 3.9 meter forward jumps under Martian gravity conditions.

-- Supplementary video: \textit{ https://www.youtube.com/watch?v=qlSJ3P87A4A}
\end{abstract}

\section{Introduction}
\vspace{-1ex}

Traditional space exploration has been dominated by the rover and lander form factor due to their historic success in returning scientific images and data when deployed on the Moon and Mars~\cite{arzo2022essential,vasavada2022mission}. Their wheeled design allows for efficient exploration of the flatter part of the lunar and planetary surfaces, but they may struggle on steep slopes, loose soil, and rough terrain with large obstacles~\cite{arm2023scientific}. However, many scientifically interesting areas are located in harder-to-reach places that rovers would struggle to get to. An important example is lava tubes, considered to bring together the trifecta of science, exploration, and resources~\cite{sauro2020lava,kolvenbach2024lunarleaper}. This motivates the use of alternative form factors that offer compelling advantages for robotic exploration~\cite{arm2023scientific}. 

Legged robots present a promising solution, having demonstrated significant improvements in capabilities and robustness in recent years~\cite{tranzatto2022cerberus}. The reduced gravity environments of Mars (\SI{3.71}{\meter\per\second\squared}), or other planetary bodies such as the Moon (\SI{1.62}{\meter\per\second\squared}), particularly favor dynamic locomotion: jumping maneuvers that would be challenging on Earth become feasible, enabling robots to overcome obstacles significantly larger than their body size~\cite{olsen2025olympus}. Some concepts propose continuous jumping as a primary mode of locomotion~\cite{Spacebok2}. However, controlling these dynamic behaviors poses significant challenges. Jumping requires precise coordination during takeoff, in-flight attitude control, and coordinated landing, all while adapting to uneven terrain and possible loose soil. The complexity of 3D attitude reorientation during flight and the uncertainty of contact schedules render classical control approaches impractical~\cite{Spacebok2,olsen2025olympus,olsen2025ral}.

\begin{figure}[t]
    \centering
    \vspace{-1ex}
    \includegraphics[width=0.68\textwidth]{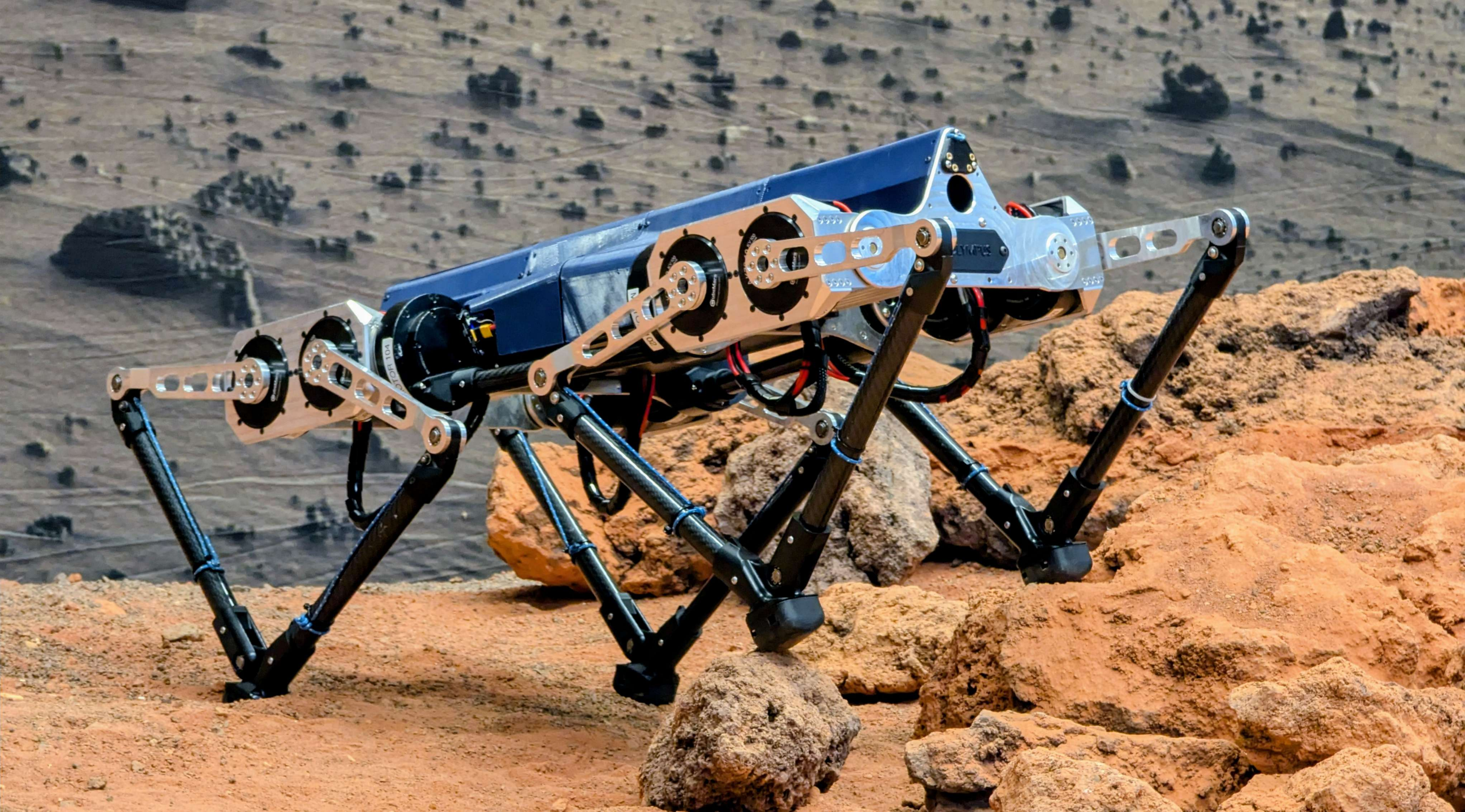}
    \vspace{-2.5ex}
    \caption{\robotname standing in a Mars analog environment.}
    \label{fig:front_figure}
    \vspace{-4ex}
\end{figure}

Building upon previous work with \robotname (Fig.~\ref{fig:front_figure}), a quadruped platform optimized for powerful jumping and in-flight attitude control in low-gravity environments~\cite{olsen2023design}, we investigate the use of deep reinforcement learning (DRL) to enable highly dynamic quadrupedal locomotion on Martian terrain. Our contributions include:
\vspace{-1ex}
\begin{itemize}
    \item A DRL-based attitude control policy capable of rapid in-flight reorientation to ensure safe landings, achieving \ang{90} rotations in \SI{2.6}{\second} during experimental validation on robot hardware.
    
    \item DRL policies for walking, vertical jumping, and forward jumping trained for Martian gravity conditions. Simulations demonstrate vertical jumps up to \SI{3.1}{\meter} and horizontal jumps up to \SI{3.9}{\meter} under Mars gravity. An investigation of optional parallel spring assistance demonstrated an increase in jump height and distance.
    
    \item A hierarchical policy deployment framework that enables traversal of obstacles significantly larger than the robot through coordinated multi-policy execution in planetary exploration scenarios.

\end{itemize}
\vspace{-1ex}

The remainder of this paper is organized as follows: Section \ref{Sec:related_work} reviews related work. Section \ref{sec:robot_system} describes the \robotname quadruped. Section \ref{sec:SimSetup} details the motor and spring models. Section \ref{sec:DRL_pipeline} outlines the training methodology. Section \ref{sec:sim_studies} shows simulation results. Section \ref{sec:experimental_validation} presents experimental validation of the attitude control policy. Section~\ref{sec:planetary_exploration_pipeline_validation} presents the planetary exploration pipeline, while Section \ref{sec:conclusion} draws conclusions.

\vspace{-1ex}

\section{Related Work}
\label{Sec:related_work}

Landers, rovers, and recently, helicopters have proven very successful for robotic space exploration~\cite{patel2023towards}. At the same time, several alternative approaches with unique robots, including legged robots, have been proposed for surface and subsurface exploration of planetary bodies~\cite{arm2023scientific,doyle2021recent}. Jumping legged robots have also been proposed, with recent work enabling such mobility using reinforcement learning controllers~\cite{Spacebok2}. One key advantage jumping legged robots have for such a task is their ability to jump over obstacles larger than themselves, especially in lower gravity, such as on Mars. Special areas of interest where traditional robots and helicopters might struggle would be Martian Lava Tubes and rough terrain at high altitude~\cite{olsen2025olympus}.


Beyond proven platforms like rovers and helicopters~\cite{mier-hicks_sample_2023}, proposed robotic systems for planetary exploration include pit-bots~\cite{thangavelautham2017flying} and legged robots~\cite{arm2023scientific,kolvenbach2024lunarleaper}. Among these emerging platforms, legged systems are particularly well-suited for complex terrain due to their discrete footholds and ability to overcome larger obstacles than traditional rovers. Achieving robust locomotion on such systems in challenging planetary exploration scenarios requires advanced control strategies.

Reinforcement learning has emerged as a powerful approach for controlling complex quadruped behaviors~\cite{tan2018sim}. Recent advances have seen extremely robust policies for walking policies~\cite{lee2020learning}, as well as progress in the tasks of jumping~\cite{olsen2025ral,atanassov2024curriculumbased}, and attitude control~\cite{Spacebok2,elflight}.

Building upon the work and ideas of \cite{Spacebok2,olsen2025olympus,olsen2025ral}, we present a set of high-performing DRL policies for walking, vertical jumping, forward jumping, and attitude control, that when combined in a hierarchical task-dependent state machine showcase the capabilities of the quadruped robot \robotname and the potential for jumping legged robots for lower gravity planetary exploration. 

\vspace{-1ex}

\section{Planetary Exploration System}
\label{sec:robot_system}

This section describes the robot, design optimization process, resulting hardware specifications, and onboard control architecture.

\subsection{Robot Design and Optimization}

\begin{figure}[t]
    \centering
    \begin{minipage}[c]{0.5\textwidth}
        \centering
        \includegraphics[width=\linewidth]{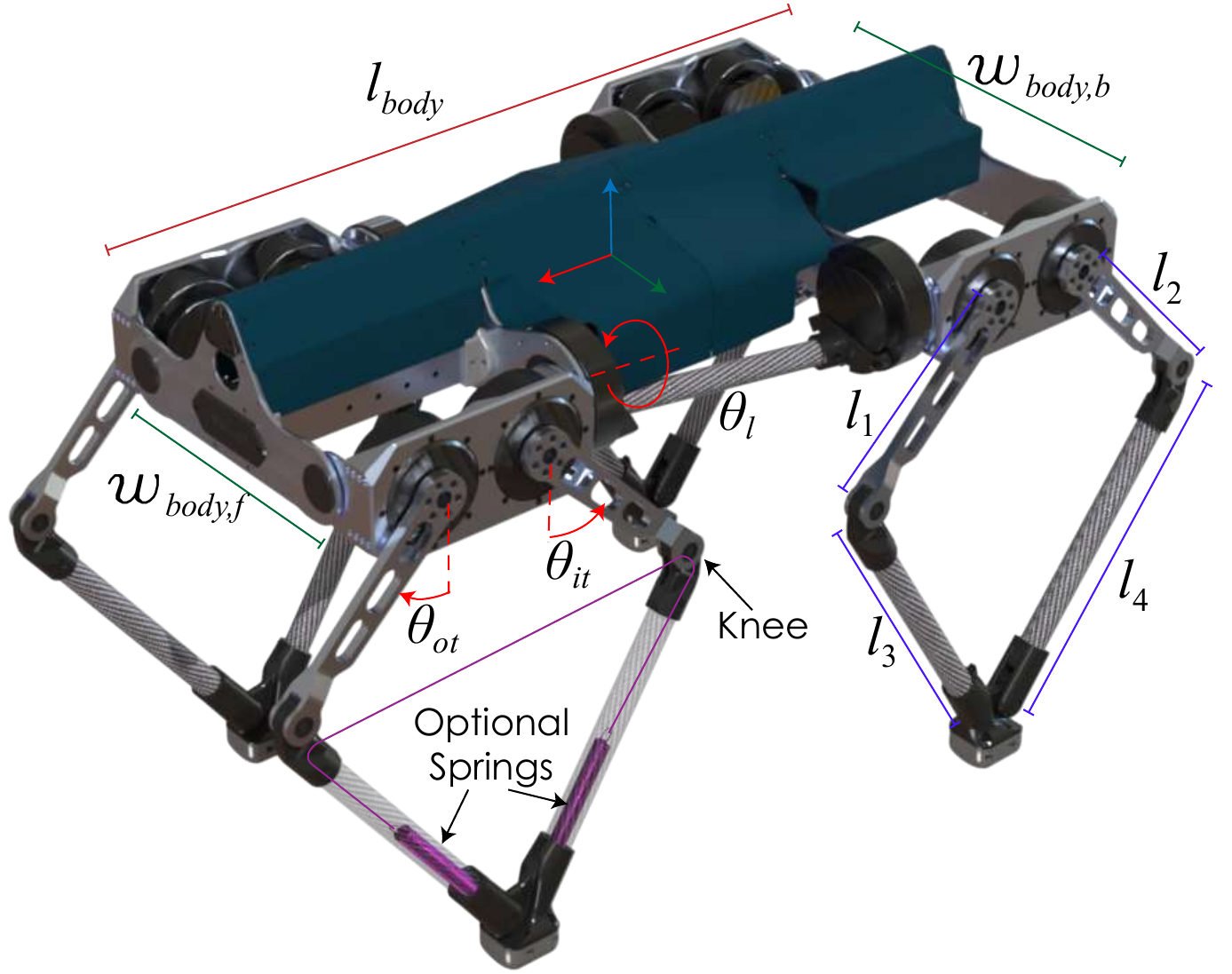}
        \caption{\robotname design with annotated body and leg parameters.}
        \label{fig:olympus_explain_figure}
    \end{minipage}\hfill
    \begin{minipage}[c]{0.5\textwidth}
        \centering
        
        \makeatletter
        \def\@captype{table}
        \makeatother
        
        \caption{Optimized Body and Leg Parameters}
        \label{tab:OlympusParameters}
        \begin{tabular}{lcc} \thickhline
        Parameter                    & Symbol         & Value\\ \hline
        Front-back leg distance      & $l_{body}$     & \SI{0.6}{\meter}\\
        Front leg separation         & $w_{body,f}$   & \SI{0.21}{\meter}\\
        Back leg separation          & $w_{body,b}$   & \SI{0.3}{\meter}\\
        Link 0 length                & $l_0$          & \SI{0.09}{\meter} \\
        Link 1 and 2 length          & $l_1$ \& $l_2$ & \SI{0.175}{\meter}\\
        Link 3 and 4 length          & $l_3$ \& $l_4$ & \SI{0.3}{\meter}\\
        Spring stiffness             & $k$            & \SI{800}{\newton\per\meter} \\ \thickhline
        \end{tabular}
    \end{minipage}
    \vspace{-4ex}
\end{figure}

The quadrupedal robot \robotname was designed for dynamic locomotion in Martian gravity environments. It employs three degrees of freedom per leg, and its 5-bar leg design provides both a large workspace for in-flight attitude control and excellent jumping capabilities. Parallel spring utilization is possible, albeit optional, to aid in jumping maneuvers. The robot was optimized to maximize jumping performance and reorientation capabilities while maintaining dynamic walking ability under Mars gravity conditions. The optimization employed a grid search over the morphological design space, varying body dimensions and leg parameters. Body dimensions searched included body length $l_{body}$, front leg separation $w_{body,f}$, and back leg separation $w_{body,b}$. The 5-bar leg parameters search space comprised link lengths $l_1$ through $l_4$ and spring stiffness $k$. The grid search prioritized vertical jump height, horizontal jump distance, and angular reorientation rates. Complete details of the optimization methodology and design space exploration are provided in~\cite{olsen2025olympus}. The final optimized parameters for the robot are listed in Table~\ref{tab:OlympusParameters} and illustrated with the robot design in Fig.~\ref{fig:olympus_explain_figure}, where for each leg $\theta_{l}$ is the lateral motor angle, $\theta_{ot}$ is the outer transversal motor angle, and $\theta_{it}$ denotes the inner transversal motor angle.


The robot has a mass of \SI{14.5}{\kilogram}, a total body length of \SI{0.67}{\meter}, and is actuated by twelve CubeMars torque-controlled brushless DC motors (AK80-9 for lateral, AK70-10 for transversal). The maximum torque for the lateral motors is $\tau_{max,l} =$ \SI{18.0}{\newton\meter} and for the transversal motors $\tau_{max,t} =$ \SI{24.8}{\newton\meter}. The 5-bar linkage leg design enables the optional utilization of integrated springs placed in the leg, connected via a cord through pulleys in the knee joints of the 5-bar leg, enabling increased energy output during jumps and adding compliance and energy storage during landing. Unless otherwise stated, all experiments use the robot configuration without springs.


\subsection{Onboard Control Architecture}

\begin{figure}
    \centering
    \includegraphics[width=0.85\textwidth]{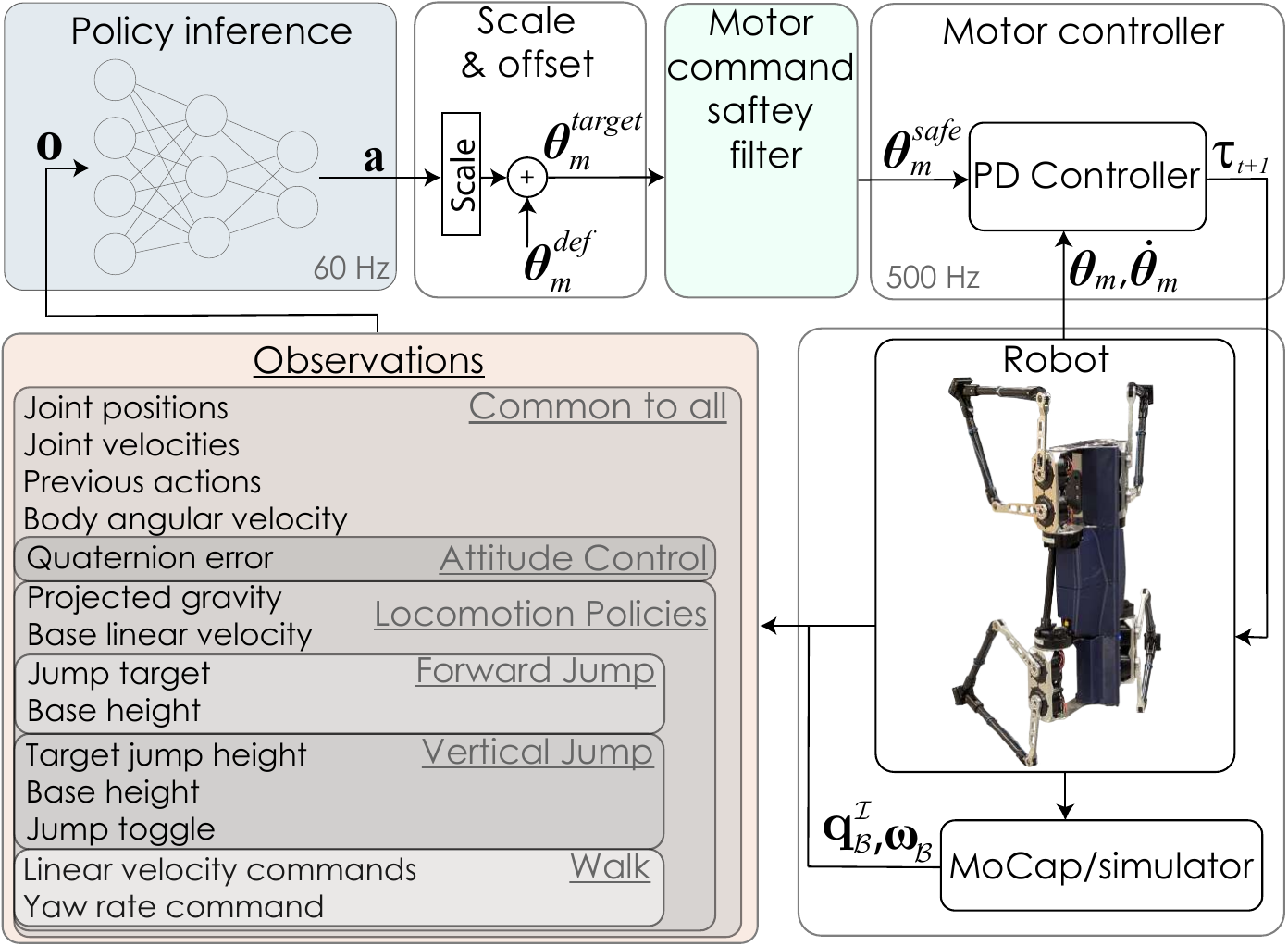}
    \vspace{-2ex}
    \caption{Controller architecture on the robot and in simulation. With common and task-specific observations.}
    \label{fig:control_loop}
    \vspace{-4ex}
\end{figure}

The deployed control architecture operates as a closed-loop system (Fig.~\ref{fig:control_loop}), where the policy receives task-specific (i.e., forward/vertical jumping, walking, attitude maneuvering) observations $\mathbf{o}$ and outputs actions $\mathbf{a}$. The actions are rescaled and offset to center around default joint angles $\theta_m^{def}$, producing motor target angles $\boldsymbol{\theta}^{target}_m$. Task-specific rescaling applied: walking uses \ang{60} for all motors, attitude control uses \ang{90} for all motors, and jumping uses \ang{15} for lateral motors and \ang{90} for transversal motors. All lateral and transversal motors have default angles of $\theta_l^{def} = \ang{0}$ and $\theta_t^{def} = \ang{45}$, respectively. To ensure safe operation, target angles are passed through a safety filter to produce safe motor targets $\boldsymbol{\theta}_m^{\text{safe}}$. These filtered angles are then tracked by PD motor controllers, which generate reference torques $\boldsymbol{\tau}_{t+1}$ commanded to the actuators. The control architecture is identical in simulation and hardware experiments. In simulation, NVIDIA Isaac Lab provides all states directly. On hardware, policy inference run at \SI{60}{\hertz} on an NVIDIA Jetson Orin NX onboard computer, receiving current body orientation  $\mathbf{q}_\mathcal{B}^{\mathcal{I}}$  and angular velocity $\boldsymbol{\omega}_{\mathcal{B}}$ from a Vicon motion capture (MoCap) system, and joint angles $\boldsymbol{\theta_m}$ and velocities $\boldsymbol{\dot{\theta}}_m$ directly from the motors, where  $\mathcal{B}$ is the body frame and  $\mathcal{I}$ is the inertial frame.

\section{Simulation Setup}
\label{sec:SimSetup}

\subsection{Motor Model}
The motor control in simulation combines a saturated PD control model with a zero-order torque-speed characteristic model. The PD controller generates desired torques based on position errors, subject to the motor's physical torque-speed limitations described by Equation~\ref{eq:motormodel}.

\vspace{-1ex}
\small
\begin{equation}
\label{eq:motormodel}
|\tau| \leq \tau_{\text{max}}(\dot{\theta}) =
\begin{cases}
\tau_{\text{max}} & \text{if } |\dot{\theta}| \leq \dot{\theta}_{\text{cut}} \\[1pt]
\tau_{\text{max}} \left(1 - \frac{|\dot{\theta}| - \dot{\theta}_{\text{cut}}}{\dot{\theta}_{\text{no-load}} - \dot{\theta}_{\text{cut}}} \right) & \text{if } \dot{\theta}_{\text{cut}} < |\dot{\theta}| < \dot{\theta}_{\text{no-load}} \\[1pt]
0 & \text{if } |\dot{\theta}| \geq \dot{\theta}_{\text{no-load}}
\end{cases}
\end{equation}
\normalsize
where $\tau$ is the applied motor torque in simulation, limited by the motor's torque-speed curve regardless of the PD controller's desired output. $\tau_{\text{max}}$ is the rated motor torque at zero speed, $\dot{\theta}$ is the joint angular velocity, $\dot{\theta}_{\text{cut}}$ is the cutoff speed below which full torque is available, and $\dot{\theta}_{\text{no-load}}$ is the no-load speed at which torque becomes zero. The model parameters are derived from manufacturer specifications, with a more conservative no-load speed applied during attitude control policy training to ensure safer movements, as little to no external forces are applied to the legs during deployment, potentially leading to extremely fast movements.

\subsection{Safety Constraints}

Three actuator constraints are enforced in simulation to aid in training: 1) Task-specific torque saturation, where $\tau_{\text{max,task}}$ is the maximum allowed torque for a given task, due to the varied maximum torque needed for each task. This is set to \SI{10}{\newton\meter} for walking, \SI{12}{\newton\meter} for attitude control, and jumping utilizes the maximum torque possible from the motors. 2) Inspired by \cite{Spacebok2}, we employ velocity-dependent braking constraints for attitude control to prevent excessive joint speeds by restricting torque application when the joint velocity magnitude $|\dot{\theta}|$ exceeds a set safe threshold $\dot{\theta}_{\text{safe}}$, thus permitting only dissipative torques opposing the direction of motion. This constraint is applied only in simulation for attitude control. 3) Joint angle limits are enforced to prevent mechanical self-collisions through hard position constraints $\boldsymbol{\theta}_{\text{min}} \leq \boldsymbol{\theta}_m \leq \boldsymbol{\theta}_{\text{max}}$, where $\boldsymbol{\theta}_{\text{min}}$ and $\boldsymbol{\theta}_{\text{max}}$ are the lower and upper joint limits, respectively. Additionally, the five-bar linkage geometry requires a coupling constraint on the transversal joints: $\theta_{\text{l}} \leq \theta_{\text{it}} + \theta_{\text{ot}} \leq \theta_{\text{u}}$, where $\theta_{\text{l}}$ and $\theta_{\text{u}}$ are the lower and upper traversal sum limits, respectively. These angle constraints are enforced both in simulation and on the robot through the safety filter mentioned in Section~\ref{sec:robot_system}.

\subsection{5-Bar Spring Model}
The \robotname quadruped can incorporate integrated parallel springs that store energy during leg compression and release it during jumping to increase jump height. In simulation, these are modeled as virtual springs connecting the knee joints of each 5-bar leg mechanism, with spring forces applied as additional torques ($\boldsymbol{\tau}_{\text{spring}}$) to the motor joints. The total joint torque ($\boldsymbol{\tau}_{\text{total}}$) combines active motor torque ($\boldsymbol{\tau}_{\text{motor}}$) with passive spring contributions:

\vspace{-1ex}
\begin{equation}
\boldsymbol{\tau}_{\text{total}} = \boldsymbol{\tau}_{\text{motor}} + \boldsymbol{\tau}_{\text{spring}}
\end{equation}


The spring model employs an extension spring with the force proportional to the extension of the knee-to-knee distance beyond its rest length, computed using the forward kinematics of the 5-bar linkage. Spring torques are obtained through the Jacobian transpose mapping from spring forces to joint torques. Due to the physical spring design~\cite{olsen2023design}, the model exhibits two-stage behavior during squat motions based on the combined angle $\theta_{\text{combined}} = \theta_{\text{it}} + \theta_{\text{ot}}$. The two-stage torque operates over different angular ranges: stage one for $\theta_{\text{combined}} \in [\ang{0}, \ang{180}]$ and stage two for 
$\theta_{\text{combined}} \in [\ang{180}, \ang{240}]$. The torque is defined as:

\begin{equation}
\boldsymbol{\tau}_{\text{spring}} = 
\begin{cases}
\mathbf{J}^T(\boldsymbol{\theta}) \mathbf{F}_{\text{spring}} & \text{if } \theta_{\text{combined}} \leq \ang{180} \\
\mathbf{F}_{\text{spring}} \mathbf{r}(\theta_{\text{combined}}) & \text{if } \theta_{\text{combined}} > \ang{180}
\end{cases}
\end{equation}

where $\mathbf{J}^T(\boldsymbol{\theta})$ is the leg Jacobian transpose mapping spring forces to joint torques, $\mathbf{F}_{\text{spring}}$ is the spring force vector, and $\mathbf{r}(\theta_{\text{combined}})$ is the angle-dependent moment arm during deep compression, where the spring attachment acts as a pulley wrapped around the motor axles. The spring mechanism is illustrated in Fig.~\ref{fig:olympus_explain_figure}.

\section{DRL Pipeline}
\label{sec:DRL_pipeline}

This section presents the reinforcement learning training methodology for the developed policies. All policies utilize reward functions composed of weighted sums of multiple terms. The walking, vertical jumping, and forward jumping policies, along with the reference state initialization scheme, build upon the framework presented in~\cite{olsen2025ral} tailored to Mars gravity operations, along with observations and full reward descriptions. The attitude controller improves upon the work of \cite{olsen2025olympus}. The different observations for each policy during training and deployment are shown in Fig.~\ref{fig:control_loop}. The rewards are listed in Table~\ref{tab:rewards_all}. All policies are also subject to regularization rewards, encouraging smoother and safer motions.

\subsection{Attitude Control} \label{sec:attitude_control}
The objective of the attitude control policy $\pi_{AC} $ is to control the orientation of the robot's base by using its legs as reaction masses and altering the robot's inertia to control the robot's attitude.  The observation vector is given by:

\vspace{-1ex}
\begin{eqnarray}
        \mathbf{o} = [\mathbf{q}_\mathcal{B}^{\mathcal{R}}~\boldsymbol{\omega}_\mathcal{B}~\boldsymbol{\theta}_{m}~\dot{\boldsymbol{\theta}}_m ~\mathbf{a}_{t-1}],
\end{eqnarray}
where the quaternion error $\mathbf{q}_\mathcal{B}^{\mathcal{R}} = (\mathbf{q}_\mathcal{R}^{\mathcal{I}})^* \otimes \mathbf{q}_\mathcal{B}^{\mathcal{I}}$ represents the relative orientation between the current body orientation $\mathbf{q}_\mathcal{B}^{\mathcal{I}}$ and the desired orientation $\mathbf{q}_\mathcal{R}^{\mathcal{I}}$ ($\mathcal{R}$ denotes the reference frame), computed using quaternion conjugate $(^*)$ and multiplication $(\otimes)$, while
$\mathbf{a}_{t-1}$ are the previous actions.

Table~\ref{tab:rewards_all} lists the reward components. The primary reward term is the \textit{Quaternion error} reward, which uses two kernels with different widths to provide fine-grained ($\sigma_1$) and coarse ($\sigma_2$) orientation feedback, where the policy is rewarded to drive the error to zero. We denote $r_{q1} = \phi_{\sigma_1}(\mathbf{q}_{\mathcal{B}}^{\mathcal{R}})$ as the activation of the narrow kernel, which is used to conditionally enable secondary rewards only near the target orientation. The \textit{Body angular velocity} reward encourages rotational motion by rewarding angular velocity aligned with the rotation axis $\boldsymbol{\phi}$ from $\mathbf{q}_{\mathcal{B}}^{\mathcal{R}}$. This reward is disabled when $\mathbf{q}_{\mathcal{B}}^{\mathcal{R}} < \ang{5}$ to allow convergence. The \textit{Stability reward} encourages low angular velocity near the target. \textit{Landing position} rewards drive lateral and transversal joints toward default angles, preparing the leg configuration for landing. Here, $\boldsymbol{\theta}_l \in \mathbb{R}^4$ and $\boldsymbol{\theta}_t \in \mathbb{R}^8$ denote the vectors of all lateral and transversal joint angles across all four legs.  The \textit{Symmetry rewards} encourage coordinated leg motion. The lateral symmetry term rewards similarity in lateral joint angles between front and back legs on each side ($\boldsymbol{\theta}_{l,FL}$ and $\boldsymbol{\theta}_{l,BL}$ for left, $\boldsymbol{\theta}_{l,FR}$ and $\boldsymbol{\theta}_{l,BR}$ for right), while the transversal symmetry term rewards similarity between all inner and outer transversal joints ($\boldsymbol{\theta}_{it}$ and $\boldsymbol{\theta}_{ot}$). Episodes are terminated upon self-collision.  Key differences from~\cite{olsen2025olympus} are the focus on faster reorientation speed, symmetry rewards  (encouraging front-back lateral matching and inner-outer transversal matching) leading to cleaner movements, and landing configurations to prepare for actual in-flight use with jumping policies. Also during deployment, a linear interpolation between default and commanded joint positions is applied when the orientation error is small (within \ang{5}) to reduce oscillation~\cite{elflight}.

\begin{table}[t]
    \caption{Reward Formulations for All Policies. Notation: $\phi_\sigma(x) := \exp(-\frac{x^2}{\sigma^2})$, $\psi_\sigma(x) := \exp(-\frac{|x|}{\sigma})$.}
    \vspace{-1ex}
    \label{tab:rewards_all}
    \centering
    \small
    \setlength{\tabcolsep}{4pt}  
    \begin{tabular}{ll|ll} \thickhline
        \multicolumn{2}{c|}{\textbf{Attitude Control}} & 
        \multicolumn{2}{c}{\textbf{Walking}} \\
        \hline
        
        Quaternion error &  $\phi_{\sigma_1}(\mathbf{q}_{\mathcal{B}}^{\mathcal{R}}) + 0.6\phi_{\sigma_2}(\mathbf{q}_{\mathcal{B}}^{\mathcal{R}})$ & 
        Linear vel. error & $\phi_{\sigma_7}(||\mathbf{v}_{xy}- \mathbf{c}_{xy}||)$ \\
        
        Body ang. vel. & $\boldsymbol{\omega}_{\mathcal{B}} \frac{\boldsymbol{\phi}}{||\boldsymbol{\phi}||} (\mathbf{q}_{\mathcal{B}}^{\mathcal{R}} \geq \ang{5})$ &
        Yaw rate & $\phi_{\sigma_8}(\omega_{z}- \omega_z^*)$ \\
        
        Stability & $\phi_{\sigma_3}(\boldsymbol{\omega}^2_{\mathcal{B}}) r_{q1}$ & 
        Vertical vel. & $v_z^2$ \\
        
        Landing lateral & $\phi_{\sigma_3}(|\boldsymbol{\theta}_{l} - \boldsymbol{\theta}_{l}^{def}|) r_{q1}$ &
        Lateral stability & $||\boldsymbol{\omega}_{xy}||^2$ \\
        
        Landing transversal & $\phi_{\sigma_4}(|\boldsymbol{\theta}_{t} - \boldsymbol{\theta}_{t}^{def}|^2) r_{q1}$ & 
        Flat & $||\mathbf{g}_{xy}||^2$ \\
        
        Symmetry lateral & $\phi_{\sigma_5}(|\boldsymbol{\theta}_{l,BL} - \boldsymbol{\theta}_{l,FL}|$ &
        Stand & $\phi_{\sigma_9}(||\boldsymbol{\theta}_{t} - \boldsymbol{\theta}_{t}^{*}||)$ \\
        
        & $+ |\boldsymbol{\theta}_{l,BR} - \boldsymbol{\theta}_{l,FR}|)$ & 
        Lateral pos. & $\phi_{\sigma_{10}}(||\boldsymbol{\theta}_{t} - \boldsymbol{\theta}_{t}^{*}||^{4})$ \\
        
        Symmetry transversal & $\phi_{\sigma_6}(|\boldsymbol{\theta}_{it} - \boldsymbol{\theta}_{ot}|)$ & 
        Transversal pos. & $\phi_{\sigma_{11}}(||\boldsymbol{\theta}_{l} - \boldsymbol{\theta}_{l}^{*}||^{10})$ \\
        
        \hline
        
        \multicolumn{2}{c|}{\textbf{Vertical Jump}} & 
        \multicolumn{2}{c}{\textbf{Forward Jump}} \\
        \hline
        
        Height & $\phi_{\sigma_{12}}(h_m - h^*)$ & 
        Tracking & $\phi_{\sigma_{18}}(|\mathbf{e}|)$ \\
        
        & $+ 3\psi_{\sigma_{13}}(h_m - h^*)$ &
        Est. tracking. & $\phi_{\sigma_{19}}(|\hat{\mathbf{e}}|)$ \\
        
        Est. height & $\phi_{\sigma_{14}}(\hat{h}_m - h^*)$ &
        & $+ 0.1\phi_{\sigma_{20}}(\hat{\mathbf{e}})$ \\
        
        & $+ 3\psi_{\sigma_{15}}(\hat{h}_m - h^*)$ &
        Symmetry & $\phi_{\sigma_{21}}(||\boldsymbol{\theta}_{t,L} - \boldsymbol{\theta}_{t,R}||)$ \\
        
        Symmetry & $\phi_{\sigma_{16}}(\text{Var}(\boldsymbol{\theta}_{t}))$ & & \\
        
        & $+ \phi_{\sigma_{17}}(||\boldsymbol{\theta}_{l}||)$ & & \\
        
        \hline
        
        \multicolumn{4}{c}{\textbf{Common Jumping Rewards}} \\
        \hline
        
        Angular vel. &  $\phi_{\sigma_{22}}(||\boldsymbol{\omega}_{\mathcal{B}}||)$ &
        Orient. error & $\phi_{\sigma_{24}}(\mathbf{q}_\mathcal{B}^{\mathcal{R}})^2$ \\
        
        Joint pos. & $\phi_{\sigma_{23}}(||\boldsymbol{\theta}_m - \boldsymbol{\theta}_m^{*}||)$ &
        Ground force & $||\mathbf{F}_{ground}||^2$ \\
        
        Soft impact & $\max(0, 1 -$ &
        Catch landing & $\text{clamp}(-v_z, 0, 1)$ \\
        
        & $ |\min(0,\frac{\mathbf{a}_{\text{body}}}{a_{\text{body,max}}} \cdot \tilde{\mathbf{v}})|)$& Damp landing & $\text{clamp}(\dot{\boldsymbol{\theta}}_{t}, 0, 1)$ \\
        
        \hline
        
        \multicolumn{4}{c}{\textbf{Regularization Rewards (All Policies)}} \\
        \hline
        
        Action rate &  $||\boldsymbol{a}^{(t)} - \boldsymbol{a}^{(t-1)}||^2$ & 
        Action clip & $||\boldsymbol{\theta}_m^{target} - \boldsymbol{\theta}_m^{safe}||^2$ \\
        
        Joint acceleration & $||\boldsymbol{\ddot{\theta}}||^2$ & 
        Joint torque & $||\boldsymbol{\tau}||^2$ \\
        
        \thickhline
    \end{tabular}
    \vspace{-4ex}
\end{table}

\subsection{Walking}
\label{sec:walking}

The walking policy's main rewards track commanded linear velocity $\mathbf{c}_{xy}$ and yaw rate $\omega_z^*$ against the robot's xy-plane velocity $\mathbf{v}_{xy}$ and angular velocity $\omega_z$. Secondary rewards penalize vertical velocity $v_z$, lateral angular velocity $\boldsymbol{\omega}_{xy}$, body tilt (via projected gravity $\mathbf{g}_{xy}$), and drive lateral and transversal joint angles toward desired angles ($\boldsymbol{\theta}_t^*$, $\boldsymbol{\theta}_l^*$). The policy is trained directly in Mars gravity using this reward structure with increased regularization and adjusted reward weights relative to the Earth gravity baseline.

\vspace{-3ex}

\subsection{Vertical Jumping}
\label{sec:vertical-jump}

The vertical jumping policy tracks the commanded jump height $h^*$ after a jump trigger is given. The policy is trained for Mars gravity with modifications to the Earth gravity reward structure. The main reward modifications include: height scaling to emphasize achieving the target height $h^*$ at larger magnitudes, wider kernel tolerance $\sigma$ reflecting the increased jump heights, and relaxed termination criteria to allow early exploration of high jumps before refining precision. The \textit{Height} reward evaluates maximum achieved height $h_{m}$ against $h^*$, while \textit{Est. height} provides continuous feedback using projectile-based estimates $\hat{h}_{m}$ during flight. \textit{Symmetry} rewards joint symmetry. Common jumping rewards encourage soft landings and penalize unwanted motion. The \textit{Soft impact} reward encourages softer landings by penalizing body acceleration $\mathbf{a}_{body}$ aligned with normalized velocity direction $\tilde{\mathbf{v}}$ when exceeding threshold $\mathbf{a_{\text{max,body}}}$. Other common rewards penalize body angular velocity, orientation error, and excessive ground forces $\mathbf{F}_{ground}$, while encouraging motor target $\boldsymbol{\theta}_m^{*}$ tracking and soft landings via clamped vertical velocity and damped joint motion.

\subsection{Forward Jumping}
\label{sec:forward-jump}
The forward jumping policy tracks a commanded target position in the xy-plane and is trained for Mars gravity with reward modifications accounting for lower gravity and extended flight times. The main reward modifications include distance scaling and wider kernel tolerances reflecting larger jump distances. The \textit{Tracking} reward encourages minimization of the horizontal tracking error $\mathbf{e}$ between current and target robot position, while \textit{Est. tracking} uses the the estimated landing error $\hat{\mathbf{e}}$ during flight, based on projectile motion. The \textit{Symmetry} reward encourages symmetric motion by penalizing differences between left and right leg transversal joint angles ($\boldsymbol{\theta}_{t,L}$ and $\boldsymbol{\theta}_{t,R}$).

\subsection{Initialization During Training}

The process of learning to perform significant jumps, where the planning horizon for the policy is much shorter than the time it takes to execute the jump and reach the target height or target landing position, is challenging. Therefore, a comprehensive curriculum-based reference state initialization strategy is used to aid in training agents to learn the correct jumping behavior. This is necessary to push the agent toward the desired state, both during the ground and flight phases of the jump. The agents are initialized in different stages of jumping maneuvers, standing, squatting, in-flight, and right before touchdown. The state of the agents in the in-flight and touchdown phase is determined using equations for projectile motion to determine where in the flight trajectory they should be, based on the desired jump height/distance. 

During training, the attitude control policy is initialized with a random orientation and zero angular momentum. Simultaneously, all motor angles are initialized across the full operational range, allowing the agent to observe all states during training.

\subsection{Sim2Real Transfer for Robot Deployment}
During training, domain randomization and noise were applied to facilitate effective Sim2Real transfer for policy deployment. Randomized physical properties included body and link masses, joint friction, motor damping, and control delays, alongside Gaussian noise on all proprioceptive observations. Accurate motor modeling via system identification was employed to further reduce the Sim2Real gap. The walking and jumping control approach builds upon prior work where policies trained with this methodology transferred successfully to hardware under Earth gravity conditions \cite{olsen2025ral}, supporting the transferability of the trained policies. The primary untested gap therefore concerns the low-gravity regime and spring integration, which cannot easily be replicated on Earth. The successful hardware transfer of the attitude control policy, trained under the same methodology, also supports the readiness of the locomotion policies.

\subsection{Neural Architectures and Implementation}
Training and simulations were conducted using IsaacLab~\cite{mittal2023orbit} with the RL Games~\cite{rl-games2021} implementation of proximal policy optimization (PPO). All policies were trained on an NVIDIA RTX 3090 GPU with parallelization across 4096 environments. Each policy employs a three-layer fully-connected multilayer perceptron (MLP) architecture. The network configurations were: attitude control [512, 256, 128], walking [256, 128, 128], vertical jumping [256, 128, 128], and horizontal jumping [128, 128, 128].

\section{Simulation Studies}
\label{sec:sim_studies}

We evaluate the trained policies in simulation under two scenarios: attitude control during free flight (zero gravity) and walking, vertical, and forward jumping under Martian gravity conditions.

\subsection{Attitude Control} 


The attitude control policy was evaluated through step response tests in free-flight conditions. Given the quaternion error $\mathbf{q}_\mathcal{B}^{\mathcal{R}}$, the policy generates actions to reach the target orientation. Convergence is defined as orientation error within \ang{5}. Three test scenarios evaluated different aspects of the reorientation capability. \textit{Single-axis response} tests applied individual \ang{90} and \ang{180} step commands to each axis (roll, pitch, yaw) independently and a \textit{3D reorientation response} test commanded a simultaneous rotation from $[\ang{-90}, \ang{90}, \ang{90}]$ (roll, pitch, yaw) to $[\ang{0}, \ang{0}, \ang{0}]$ to evaluate coordinated multi-axis control.

\begin{figure*}[!t]
    \centering
    \subfloat[][Roll, pitch, and yaw response in \ang{90} tests.]{\includegraphics[width=0.325\linewidth]{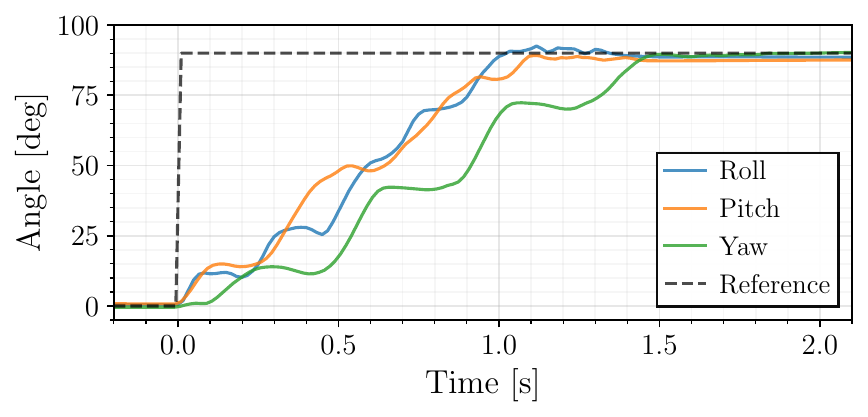}\label{fig:simulation_all_angles_together_90deg}}
    \hspace{-0ex}
    \subfloat[][Roll, pitch, and yaw response in \ang{180} tests.]{\includegraphics[width=0.325\linewidth]{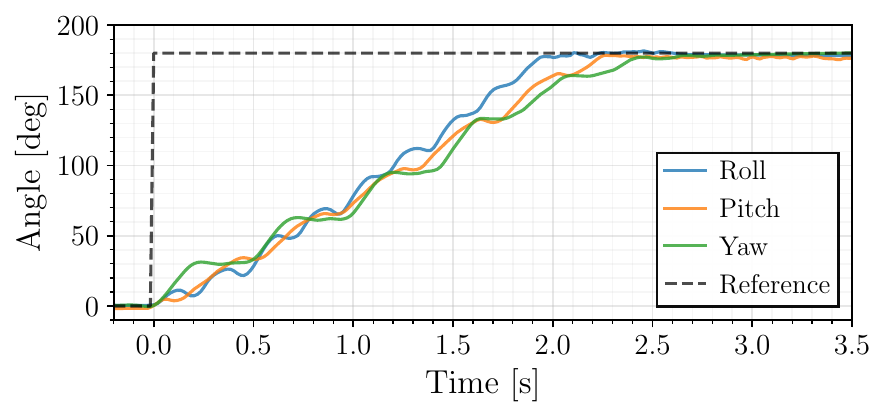}\label{fig:simulation_all_angles_together_180deg}}
    \hspace{-0ex}
    \subfloat[][3D reorientation in roll, pitch and yaw.]{\includegraphics[width=0.325\linewidth]{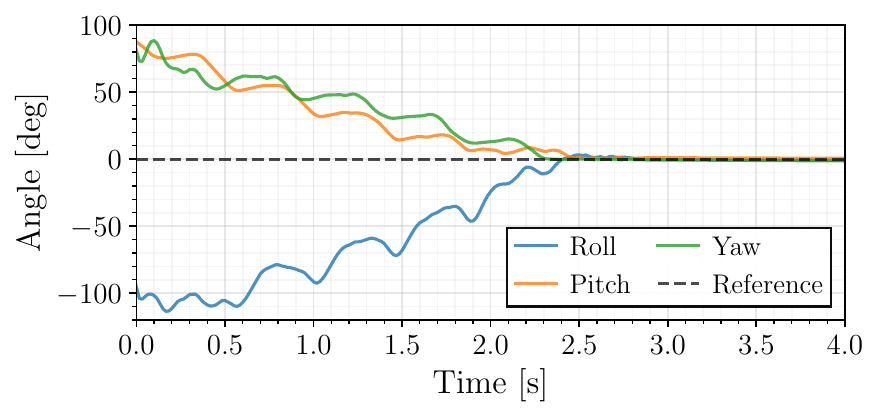}\label{fig:simulation_all_angles_composit}}
    \caption{Roll, pitch, and yaw response to a changing target orientation.}
    \label{fig:attitude_control_sim}
    \vspace{-4ex}
\end{figure*}

Fig.~\ref{fig:simulation_all_angles_together_90deg} shows the policy's response to \ang{90} single-axis commands. Roll reaches the target threshold in \SI{0.96}{\second}, pitch in \SI{1.08}{\second}, and yaw in \SI{1.45}{\second}.  The policy exhibits smooth, single-direction convergence without significant overshoot. Fig.~\ref{fig:simulation_all_angles_together_180deg} shows the response to larger \ang{180} commands, where the roll axis demonstrates the fastest response, reaching the threshold in  \SI{1.9}{\second}. Fig.~\ref{fig:simulation_all_angles_composit} presents the complex 3D maneuver results where the policy reaches the target orientation within \SI{2.45}{\second}.

\subsection{Vertical Jumping in Low Gravity}

The vertical jumping policy was evaluated through 244 jump trials with target heights spanning \SIrange{1.0}{3.5}{\meter}. The policy was trained on target heights between \SI{1.8}{\meter} and \SI{2.8}{\meter}. This range was set between the maximum achievable height (\SI{3.2}{\meter}) and a minimum useful height for vertical jumps. Success is defined as achieving apogee height within \SI{0.2}{\meter} of the commanded target height. Fig.~\ref{fig:vertical_no_spring} demonstrates strong height tracking performance across the tested range. The policy achieves a mean absolute error of \SI{0.123}{\meter}, an overall success rate of 88.9\%, and a max jump height of \SI{3.1}{\meter}. Notably, all non-successful attempts (apogee errors exceeding \SI{0.2}{\meter}) occurred for targets above \SI{2.95}{\meter}, which lies outside the training distribution. Within the trained range of \SIrange{1.8}{2.5}{\meter}, the policy demonstrates high reliability, while extrapolation to higher targets shows some degradation.

\subsection{Forward Jumping in Low Gravity}
\begin{figure}[t]
    \centering
    \subfloat[][Vertical jumps]{\includegraphics[width=0.4\linewidth]{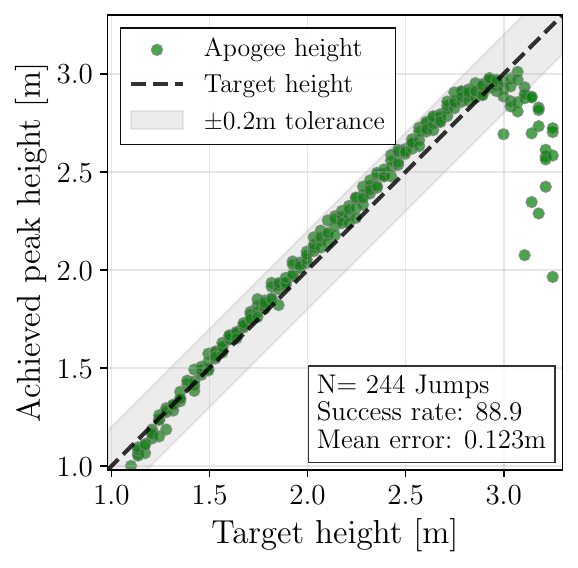}\label{fig:vertical_no_spring}}
    \hspace{-1ex}
    \subfloat[][Forward jumps]{\includegraphics[width=0.38\linewidth]{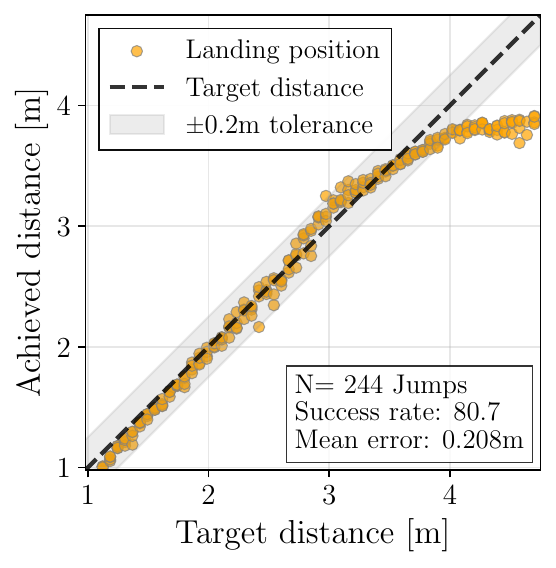}\label{fig:Forward_no_spring}}
    \caption{Jump target vs achieved jumps without spring.}
    \label{fig:jump_}
    \vspace{-4ex}
\end{figure}

The forward jumping policy was evaluated through 244 forward jumps with targets ranging from \SIrange{1.0}{4.5}{\meter} under Mars gravity conditions. The training range of the policy was between \num{1.5} and \SI{4.1}{\meter}.  Targets above \SI{4.1}{\meter} approach the extreme limit of what is achievable with the current actuator settings. Achieving the final landing position within \SI{0.2}{\meter} of the commanded target is defined as a successful jump. The policy demonstrates strong target-tracking performance across the tested range (Fig.~\ref{fig:Forward_no_spring}). The policy achieves a mean absolute error of \SI{0.208}{\meter}, a max forward jump of \SI{3.9}{\meter}, and an overall success rate of $80.7$\%. Note that all failed attempts, where landing errors exceeded \SI{0.2}{\meter}, occurred for targets above \SI{4.1}{\meter}, outside the training distribution. Within the range of \SIrange{1.0}{3.8}{\meter}, the policy demonstrates very high reliability. The policy also maintained successful landings across body orientations at touchdown up to \ang{45} roll, \ang{60} pitch, and \ang{90} yaw in simulation testing for forward jumps.

\subsection{Spring integration for Jumping}
\begin{figure}[t]
    \centering
    \subfloat[][Vertical jumps.]{\includegraphics[width=0.38\linewidth]{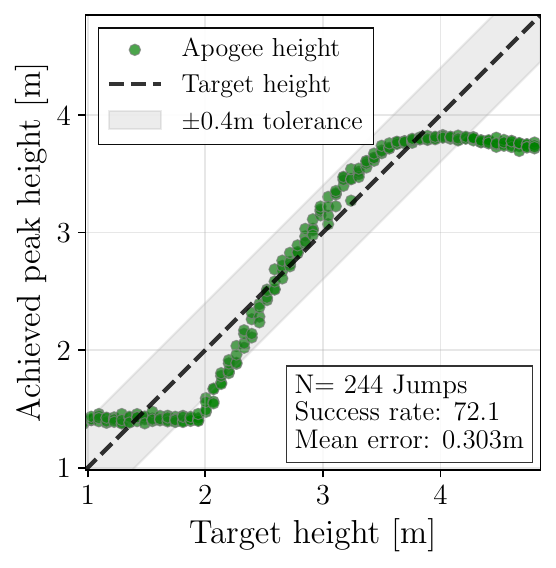}\label{fig:vertical_spring}}
    \hspace{-1ex}
    \subfloat[][Forward jumps.]{\includegraphics[width=0.38\linewidth]{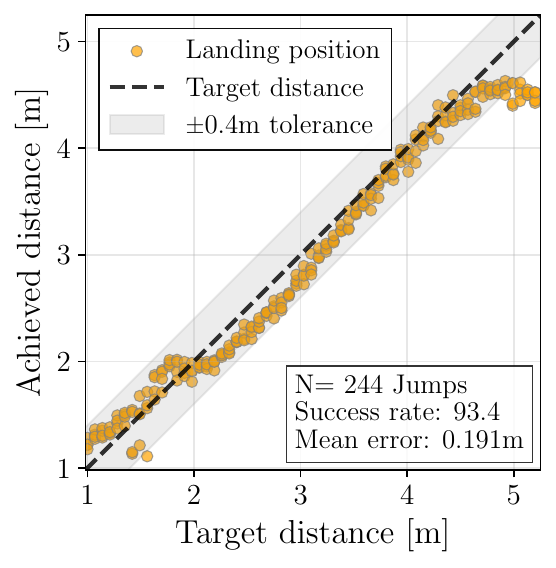}\label{fig:Forward_spring}}
    \caption{Jump target vs achieved jumps with simulated springs.}
    \label{fig:jump_with_spring}
    \vspace{-4ex}
\end{figure}

The robot design allows for integrated springs as described in Section~\ref{sec:robot_system}. To investigate the effect of having parallel springs enabled during training and deployment of the jumping policies, we trained vertical and forward jumping policies with springs enabled and a spring stiffness of \SI{800}{\newton\per\meter}, based on the optimization described in Section \ref{sec:robot_system}. The spring-enabled jump evaluation can be seen in Figure~\ref{fig:jump_with_spring}. The spring added approximately 21\% to the jump height and distance while maintaining reasonable tracking performance, showing the potential for even more powerful jumps, albeit at some tracking performance cost.

\begin{figure*}[!t]
    \centering
    \subfloat[][Roll, pitch, and yaw response in \ang{90} tests.]{\includegraphics[width=0.32\linewidth]{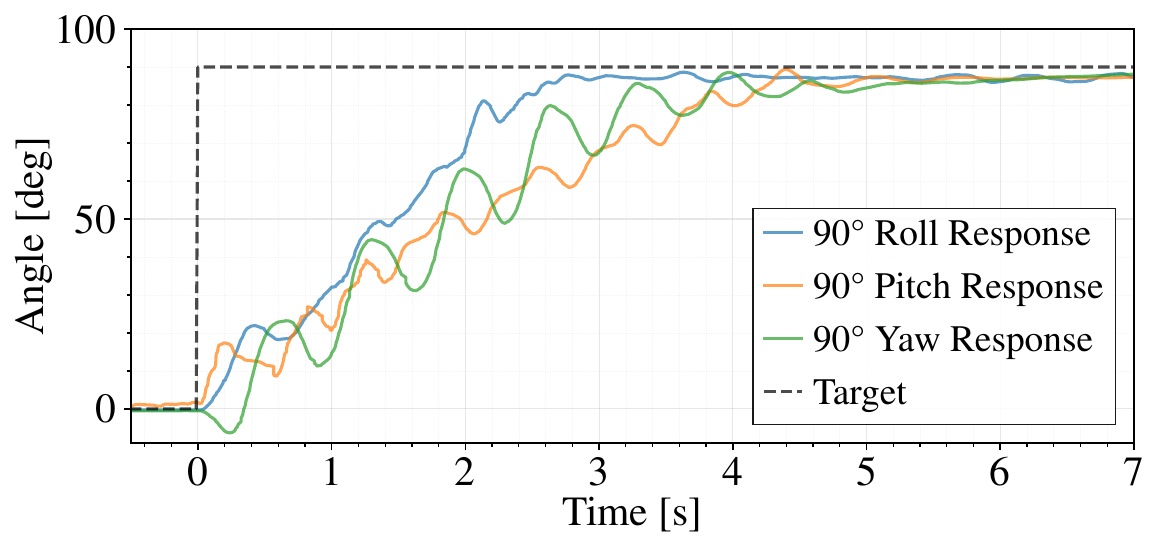}\label{fig:90_deg_test}}
    \hspace{-0ex}
    \subfloat[][Roll, pitch, and yaw response in \ang{180} tests.]{\includegraphics[width=0.32\linewidth]{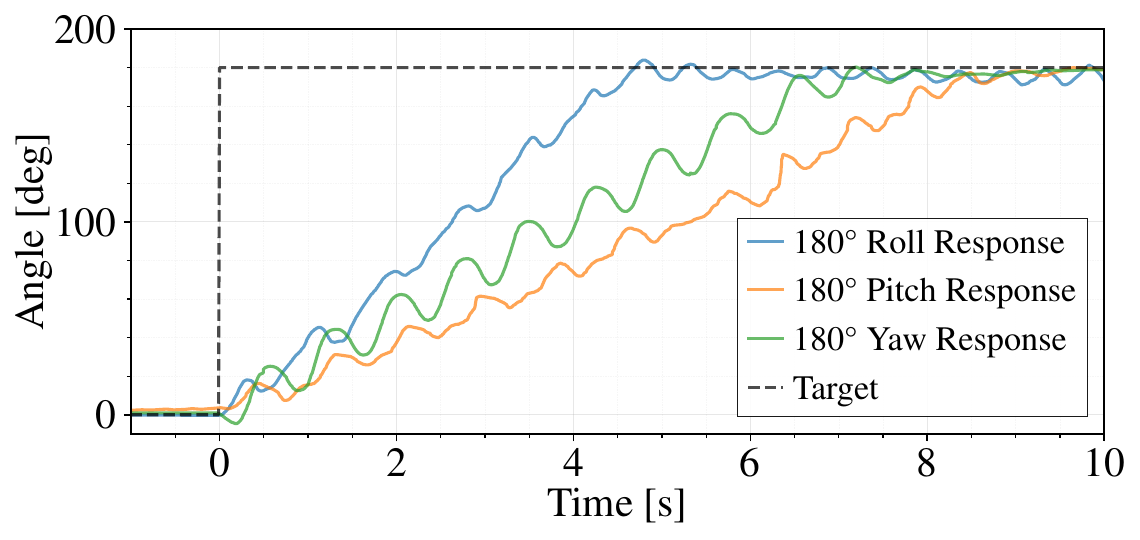}\label{fig:180_deg_test}}
    \hspace{-0ex}
    \subfloat[][Multiple step changes in target orientation.]{\includegraphics[width=0.32\linewidth]{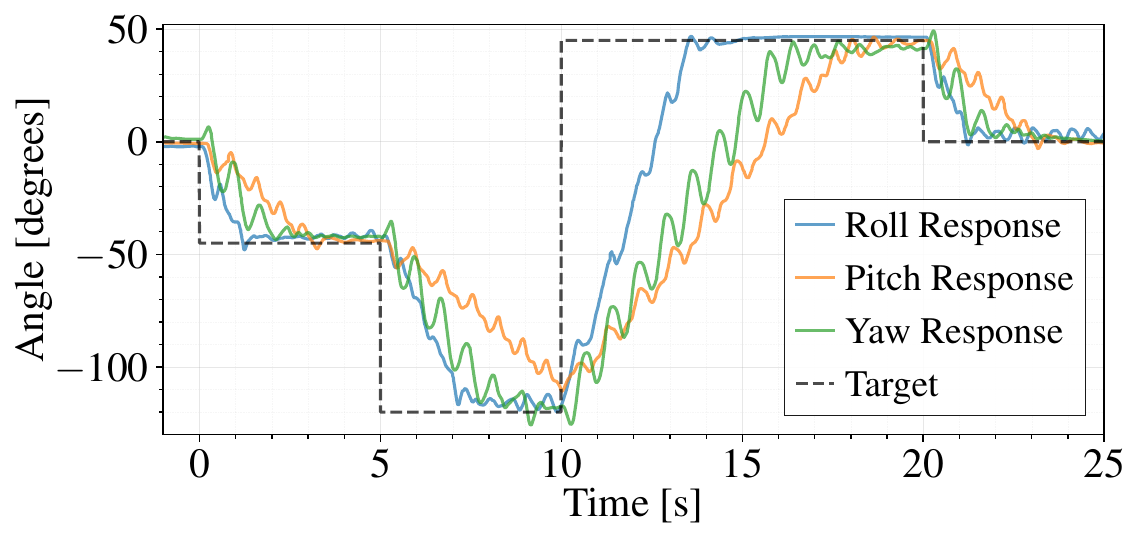}\label{fig:steps_deg_test}}
    \vspace{-0ex}
    \caption{Experimental roll, pitch, and yaw response to a changing target orientation during hardware tests.}
    \label{fig:180_deg_all}
    \vspace{-3ex}
\end{figure*}


\section{Experimental Validation}
\label{sec:experimental_validation}

This section presents experimental validation of the attitude control policy on the \robotname quadruped. State estimation is provided by a motion capture system. Motor torques are limited to \SI{12}{\newton\meter}, matching the simulation training conditions and ensuring safe operation. The tests were conducted at ESA's Orbital Robotics Lab at ESTEC.

\subsection{Attitude Control - Single Axis Rotation}
\begin{figure}
    \centering
    \includegraphics[width=0.97\linewidth]{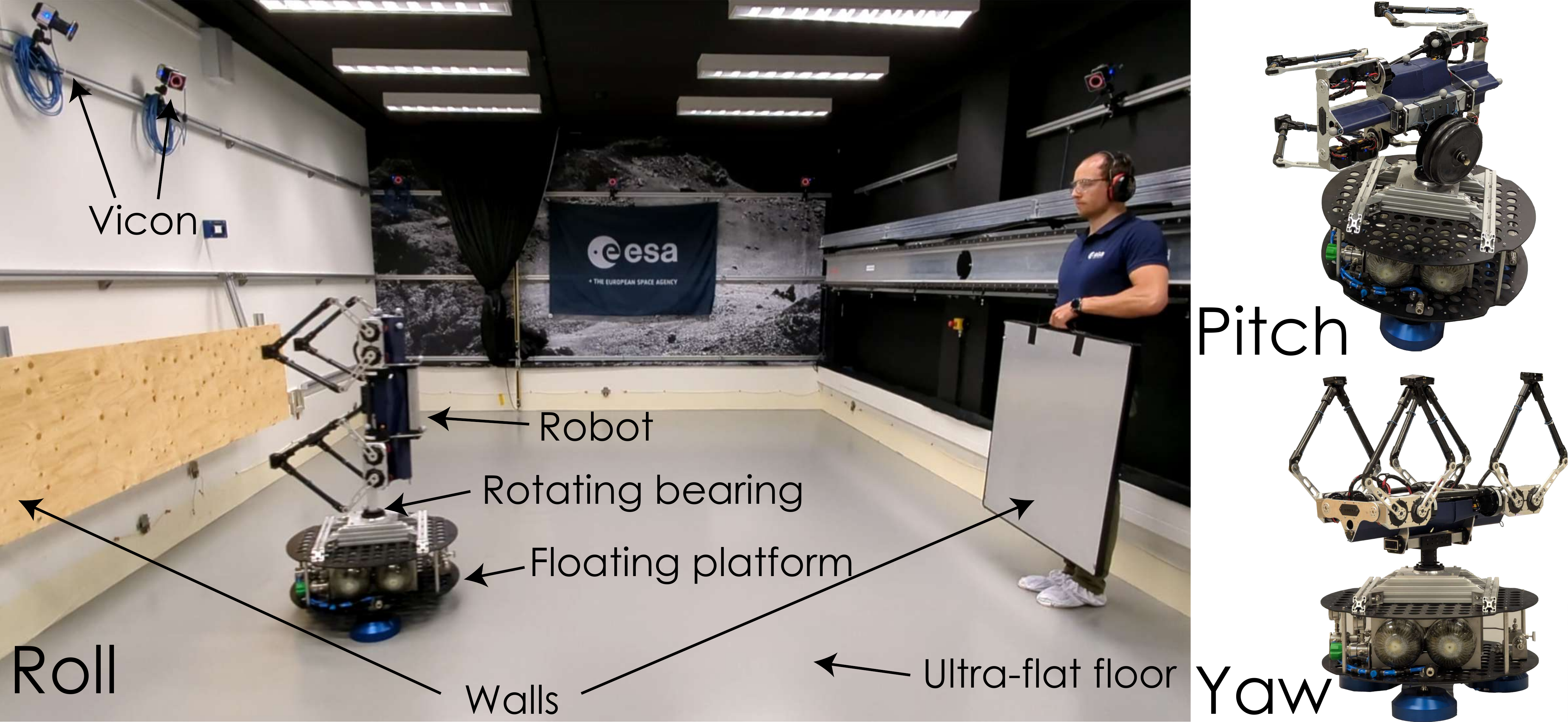}
    \vspace{-2ex}
    \caption{Attitude control test setup with floating platform on flat floor, and robot mounted on rotating rod. Showing in roll, pitch, and yaw configuration.}
    \label{fig:flat_floor_mounting}
    \vspace{-1ex}
\end{figure}

To validate the attitude control policy's reorientation capabilities, we employed a custom test stand that mounts the robot on a rotating rod, which is in turn mounted to a floating air-bearing platform. This configuration enables isolated roll, pitch, and yaw control testing by constraining motion to a single rotational degree of freedom while simulating free-flight dynamics for one axis at a time. Figure~\ref{fig:flat_floor_mounting} illustrates the robot mounting configurations for each orientation axis. Due to mechanical constraints during pitch testing, where the platform restricts leg movement, we developed a specialized policy variant $\pi_{AC\text{-pitch}}$ with constrained hip motion (limited to $\pm$ \ang{5}) to prevent leg-platform collisions.

Three reorientation tests were executed for each axis: \ang{90} step responses, \ang{180} step responses, and multi-step sequences. To ensure safe operation, given the mounting platform instability, leg motion speeds were reduced using a moving average filter (20-sample window). The resulting reorientation times are summarized in Table~\ref{tab:reorientation_times}. Note that real-world times are slower than in simulation due to this safety constraint; simulations with equivalent filtering show comparable performance, indicating the policy is capable of faster responses when hardware constraints are removed.

\begin{table}
    \centering
    \caption{Attitude Control Reorientation Times}
    \vspace{-2ex}
    \label{tab:reorientation_times}
    \setlength{\tabcolsep}{6pt}
    \begin{tabular}{lcccccc} 
    \thickhline
    \multirow{2}{*}{Test} & \multicolumn{2}{c}{Roll} & \multicolumn{2}{c}{Pitch} & \multicolumn{2}{c}{Yaw} \\
    & Sim & Real & Sim & Real & Sim & Real \\ \hline
    \ang{90} & \SI{0.96
    }{\second} & \SI{2.6}{\second} &\SI{1.08}{\second} &\SI{4.2}{\second} & \SI{1.44}{\second} & \SI{3.9}{\second} \\
    \ang{180} & \SI{1.9}{\second} & \SI{4.6}{\second} & \SI{2.3}{\second} & \SI{8.4}{\second} & \SI{2.4}{\second} & \SI{7.1}{\second} \\ \thickhline
    \end{tabular}
    \vspace{-2ex}
\end{table}

Fig~\ref{fig:90_deg_test} shows the \ang{90} step response test, with roll reaching the target orientation in \SI{2.6}{\second}.   Fig.~\ref{fig:180_deg_test} shows the \ang{180} test, 
with roll being fastest at reaching the target orientation in \SI{4.6}{\second}. Figure~\ref{fig:steps_deg_test} shows the multi-step sequence results. 



\subsection{Attitude Control - Wall Bounce}
To demonstrate integrated reorientation capabilities during dynamic maneuvers, we conducted wall-bounce experiments combining attitude control with free-floating flight phases. The robot begins mounted on the free-floating platform with an initial velocity directed toward a wall while the attitude control policy maintains feet-forward orientation. Upon reaching \SI{1}{\meter} from the wall, the robot executes a pre-programmed push-off maneuver to reverse its trajectory toward the opposite wall. Simultaneously, the attitude control policy receives a \ang{180} reorientation command to maintain feet-forward orientation for the approaching wall. This creates a continuous bounce sequence where the robot must reorient during each free-floating flight phase to achieve proper landing orientation.

\begin{figure}
    \centering
    \includegraphics[width=0.95\linewidth]{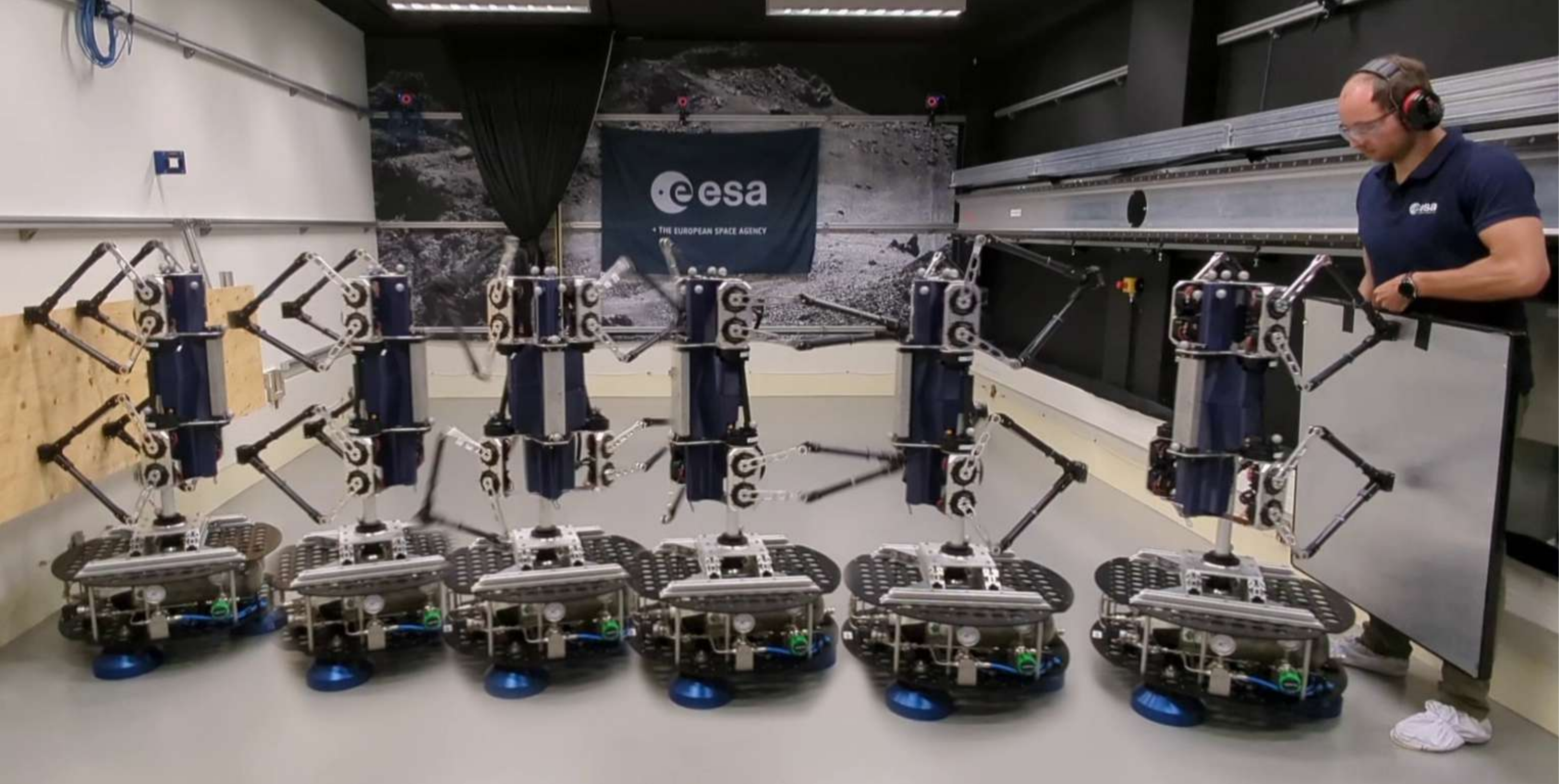}
    \vspace{-2.5ex}
    \caption{Wall bounce experimental validation of attitude control policy during dynamic maneuvers. The robot demonstrates successful \ang{180} reorientation capabilities for roll while transitioning between wall interactions.}
    \label{fig:wall_bounce}
    \vspace{-3ex}
\end{figure}

The experiment was conducted for each rotational axis (roll, pitch, and yaw) with at least five complete bounce cycles per test. Fig.~\ref{fig:wall_bounce} illustrates a representative roll sequence showing the left-to-right trajectory with successful reorientation. This test validates the policy's ability to perform rapid attitude corrections under dynamic conditions representative of free-flight scenarios in planetary exploration missions.

\section{Planetary Exploration Pipeline Testing}
\label{sec:planetary_exploration_pipeline_validation}

\vspace{-2ex}
To validate the integrated capabilities of all trained policies, we implemented a comprehensive exploration mission in simulated Martian terrain and gravity. The mission demonstrates coordinated deployment of walking, vertical jumping, horizontal jumping, and in-flight attitude control policies to traverse challenging terrain features representative of a challenging planetary exploration scenario on Mars.

A waypoint-based hierarchical controller coordinates task execution, where each waypoint specifies target position ($x, y$), orientation (yaw), and locomotion mode (walk, vertical jump, or forward jump with associated parameters such as jump distance or target height). The controller tracks position and orientation errors in the robot body frame, commanding the walking policy to approach waypoints within a \SI{0.07}{\meter} threshold. The walking policy demonstrates robust terrain traversal over obstacles up to \SI{0.15}{\meter} and inclines up to \ang{20}, while maintaining the ability to remain fully stationary under zero velocity commands, critical for consistent handover to jumping policies. Upon reaching jump waypoints, the system transitions through a predefined state sequence: the robot stabilizes in stance for \SI{1}{\second}, switches to the appropriate jumping policy for takeoff, then once the feet lose ground contact and altitude exceeds \SI{0.6}{\meter} with positive vertical velocity, the attitude control policy engages to regulate robot's body orientation to ensure safe landings. As the robot descends below \SI{0.9}{\meter}, control returns to the jumping policy for landing. After ground contact and a brief recovery period, the walking policy resumes and demonstrates recovery control capabilities, maintaining stability during jump-to-walk transitions despite momentum carry-over from aerial phases.

\begin{figure}
    \centering
    \includegraphics[width=0.999\linewidth]{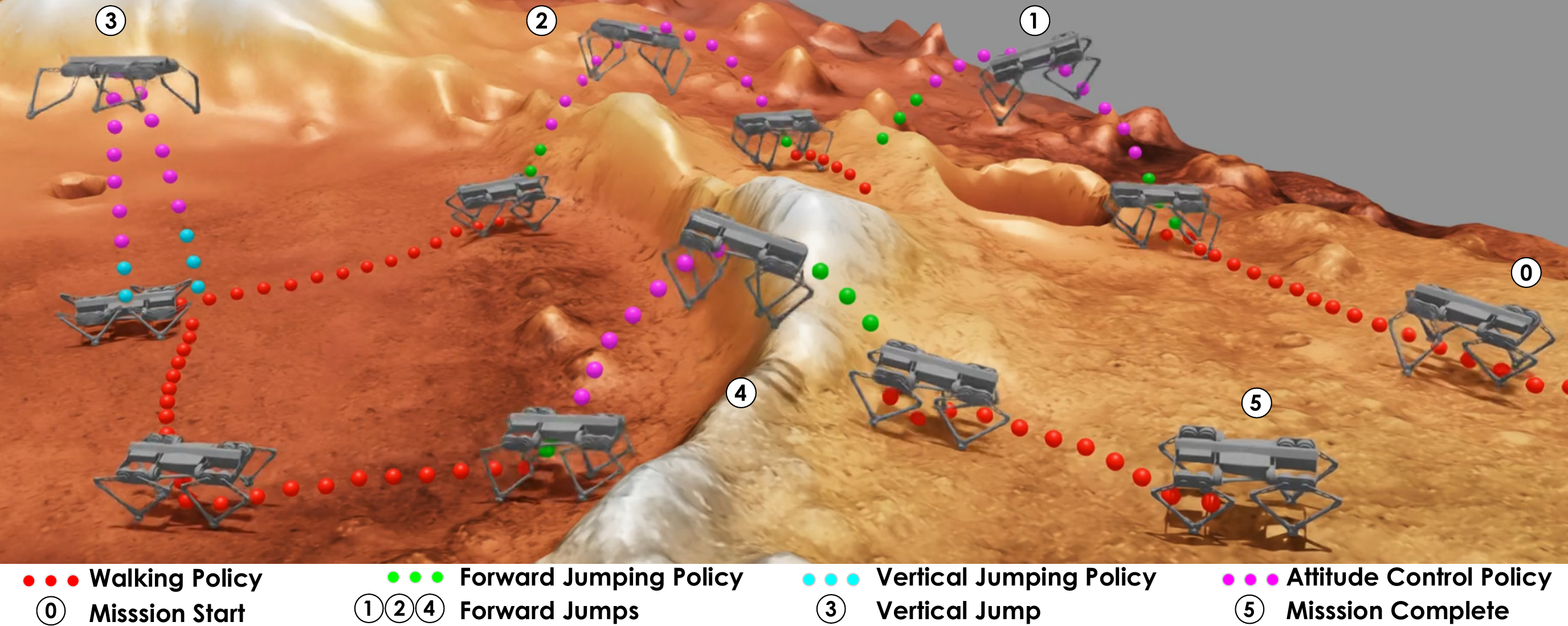}
    \vspace{-4.5ex}
    \caption{Sequential frames from integrated planetary exploration mission simulation in Isaac Lab in a single run. The color of the dots represents the task-specific policy used at that instance in the simulation. The robot successfully navigates rough terrain, jumping over a \SI{2.1}{\meter} wide crater, executes a \SI{3.5}{\meter} forward jump from a \SI{1.1}{\meter} ledge, performs a \SI{2.6}{\meter} vertical reconnaissance jump, and clears a \SI{1.1}{\meter} ledge during a forward jump. The attitude control policy maintains the desired orientation during all flight phases to ensure safe landings. Model of Mars terrain from \textit{https://sketchfab.com/gaiastucky}}
    \label{fig:planetary}
    \vspace{-3ex}   
\end{figure}

The mission incorporates multiple locomotion challenges: traversing rough terrain, jumping over a \SI{2.1}{\meter} wide crater, executing a \SI{3.5}{\meter} forward jump from a \SI{1.1}{\meter} ledge, performing a \SI{2.6}{\meter} vertical jump for elevated observation, and clearing a \SI{1.1}{\meter} ledge with challenging landing conditions. Figure~\ref{fig:planetary} presents sequential frames from a continuous simulation run demonstrating successful mission completion. The integrated system enables traversal of terrain that would be impossible using traditional exploration robots, validating the jumping legged robot multi-policy locomotion approach for planetary exploration applications.

\section{Conclusion}
\label{sec:conclusion}

\vspace{-1ex}
This work presented a reinforcement learning approach for dynamic quadrupedal locomotion in planetary exploration scenarios, demonstrating attitude control, walking, and jumping policies trained for Martian gravity and in-flight conditions. The attitude control policy achieved a reorientation time of \SI{2.6}{\second} for a \ang{90} change in orientation during hardware experiments and \SI{0.96}{\second} in simulation, while jumping tests in simulation achieved vertical jumps up to \SI{3.1}{\meter} and horizontal jumps up to \SI{3.9}{\meter}. The capabilities of the combined multi-policy planetary exploration pipeline were demonstrated by successfully traversing challenging terrain features in simulation that would be impossible with a rover. Key limitations include the lack of hardware validation for the locomotion policies under low-gravity conditions and the untested spring integration, both of which represent natural directions for future work. These results nonetheless demonstrate the viability of jumping legged robots for planetary exploration.

\bibliographystyle{ieeetr}
\bibliography{./isrr2026.bib}

\end{document}